\documentclass{sig-alternate}
\usepackage{amsmath}
\usepackage{multicol}
\begin{document}
\title{Efficient Discovery of Large Synchronous Events in Neural Spike
Streams
}

\numberofauthors{3}

\author{
\alignauthor
V.~Raajay\\
       \affaddr{Indian Institute of Science}\\
       \affaddr{Bangalore, India. 560025}\\
       \email{raajay.v@gmail.com}
\alignauthor
P.~S.~Sastry\\
       \affaddr{Indian Institute of Science}\\
       \affaddr{Bangalore, India. 560025}\\
       \email{sastry@ee.iisc.ernet.in}
\alignauthor
K.~P.~Unnikrishnan\\
       \affaddr{University of Michigan}\\
       \affaddr{Ann Arbor, USA.}\\
       \email{kpuk@umich.edu}
}    
\date{3 June, 2010}
\maketitle
\begin{abstract}
We address the problem of finding patterns from multi-neuronal
spike trains that give us insights into the multi-neuronal codes used in
the brain and help us design better brain computer interfaces. We focus on
the synchronous firings of groups of neurons as these have been shown to
play a major role in coding and communication (\cite{Grun2009}). With large electrode arrays,
it is now possible to simultaneously record the spiking activity of
hundreds of neurons over large periods of time. Recently, techniques have
been developed to efficiently count the frequency of synchronous firing
patterns. However, when the number of neurons being observed grows they
suffer from the combinatorial explosion in the number of possible patterns
and do not scale well. In this paper, we present a temporal data mining
scheme that overcomes many of these problems. It generates a set of
candidate patterns from frequent patterns of smaller size; all possible
patterns are not counted. Also we count only a certain well defined subset
of occurrences and this makes the process more efficient. We highlight the
computational advantage that this approach offers over the existing methods
through simulations.

We also propose methods for assessing the statistical significance of the
discovered patterns. We detect only those patterns that repeat often enough
to be significant and thus be able to automatically fix the threshold for
the data-mining application. Finally we discuss the usefulness of these
methods for brain computer interfaces (\cite{Donoghue2008,Coleman}).

\end{abstract}


\section{Introduction}
\label{sec:introduction}
Neurons form the basic computing elements of brain and hence, gaining the understanding of the coordinated behavior 
of groups of neurons is essential for gaining a principled understanding of the brain function. Thus, one of the
 important problems in neuroscience is that of understanding the functioning of a neural tissue 
 in terms of interactions among its neurons .

The neurons communicate with one another by means of voltage fluctuations called action potential or {\em spikes}. 
We can study the activity of a specific neural tissue by gathering data in the form of sequences of action potentials 
or spikes generated by each of a group of potentially interconnected neurons. Recent techniques, like Micro Electrode Array
(MEA), imaging of ionic concentrations etc., have enabled us in recording the activity of hundreds of neurons simultaneously.
Such recorded data, known as multi-neuronal spike train data, is a mixture of the stochastic spiking of activities of
 individual neurons as well as correlated spiking activity due to interactions or connections among neurons.
 
One way to find out the interactions among neurons is to find patterns from the spike train data(\cite{Brown2004}).
 The patterns help in understanding the relation between the spiking times of neurons which in turn can throw 
light on the interaction among neurons performing a specific function.

Various algorrithms have been developed to find interesting patterns in spike train data.
All the algorithms essentially find {\em frequent} (or less-frequent) occurences of specific patterns and try to
establish the significance of their occurence in the data, that is, statistically show that these patterns
have not occurred by chance and have occurred because of interaction among the constituent neurons.

Predominantly, two kinds of patterns have been explored, namely, 1. Sequential firing patterns 2. Synchronous firing patterns.
Sequential firing patterns are used to represent a chain of neurons firing one after the other after a certain amount of
delay in time. Such patterns have been found in data recorded from the hippocampus circuit.

Synchronous firing patterns, on the other hand, represent a group of neurons firing very close to each other in time.
The difference in the their spiking times is in the order of milliseconds. Some algorithms (\cite{Grun2002}) use 
time binning technique to find such patterns. The spiking times of neurons are binned into time intervals equal to 
time span of interest. Then the occurence of every possible pattern is checked in each time bin. Such binning techniques 
affect the time resolution of the spike times. Also some patterns that occur across time bins will be missed. 
Recently techniques \cite{Pipa2008} have been developed that avoid time-binning and count the frequency of patterns more
efficiently. However, these algorithms are all essentially correlation based and also count the frequency of all
possible patterns. When the number of neurons being observed grows they suffer from the combinatorial explosion
in the number of possible patterns and do not scale well. 

In this paper, we view this problem of finding synchronous patterns from a temporal datamining perspective. The patterns 
are represented as parallel episodes (with expiry times) discussed in the frequent episode discovery framework \cite{PSU2008}.
We use the parallel mining algorithm to discover frequent patterns whose frequency in the data is above a user specified threshold.
The algorithm is apriori based. It generates a set of candidate patterns from frequent patterns of smaller size. 
All possible patterns are not counted. Also we count only a certain well defined subset of occurrences and this makes the 
process more efficient. We highlight the computational advantage that this approach offers over the existing 
methods through simulations.

We also develop statistical techniques to establish the significance of the discovered patterns. 

The rest of the paper is organised as follows. Section~\ref{sec:parallel-episodes} describes the parallel episode 
mining algorithm that we use to discover synchronous patterns. Section~\ref{sec:significance} presents a significance
test for the parallel episodes. In Section~\ref{sec:simulations} we present simulation results that shows the 
effectiveness of our method. Concluding remarks are provided in Section~\ref{sec:conclusions}.

\section{Frequent Episode Framework for discovery of synchronous patterns}
\label{sec:parallel-episodes}

Temporal datamining is concerned with analyzing symbolic time series data to 
discover `interesting' patterns of temporal 
dependencies (\cite{Srivats-survey2005,Morchen2007}). Recently we have proposed 
that some datamining techniques, based on the so called frequent episodes framework, 
are well suited for analyzing multi-neuronal spike train data \cite{PSU2008,Diekman2009,SU08}. 
Patterns of interest in spike data such as synchronous firings by groups 
of neurons, the sequential patterns, 
and synfire chains which are a combination of synchrony and ordered firings, can be 
efficiently discovered from the data using these datamining techniques. In this 
section we first briefly outline the frequent episodes framework and then 
qualitatively describe this datamining technique for discovering frequently 
occurring synchronous patterns.

In the frequent episodes framework of temporal datamining. 
 the  data to be analyzed is a sequence of events
denoted by $\langle(E_{1},t_{1}),(E_{2},t_{2}),\ldots\rangle$ where $E_{i}$
represents an \textit{event type} and $t_{i}$ the \textit{time of occurrence} of
the $i^{th}$ event. $E_i$'s are drawn from a finite set of event types, $\zeta$.
The sequence is ordered with respect to time of occurrences of the events so
that, $t_i\le t_{i+1}$,  $\forall i$. The following is an
example event sequence containing 11 events with 5 event types.

\begin{equation}
\begin{split}
\footnotesize
\langle(A,1),(B,3),(D,5),(A,5),(C,6),(A,10), \\
(E,15),(B,15),(B,17),(C,18),(C,19)\rangle
\label{eq:data-seq}
\end{split}
\end{equation}

A \textit{parallel episode} is an ordered tuple of event types. For example,
$(A ~ B ~ C)$ is a {\em 3-node} parallel episode. 
  Such an episode is said to
\textit{occur} in an event sequence if there are  corresponding events in the data sequence.
In sequence (\ref{eq:data-seq}), the events
\{${(A,1),(B,3),(C,6)}$\} and \{$(B,3), (C,6), (A,10)$\} constitute an occurrence of the
parallel episode $(A~ B ~ C)$ .
We note here that occurrence of an episode does not
require the associated event types to occur consecutively;
there can be other intervening events between them.

The objective in frequent episode discovery is to detect {\em all} frequent episodes 
(of different lengths) from the data. 
A {\em frequent episode} is one whose {\em frequency} exceeds a 
 (user specified) {\em frequency threshold}.
The frequency of an episode can be defined in many ways. 
It is intended
to capture some measure of how often an episode occurs in an event
sequence. One chooses a measure of frequency so that frequent episode discovery is
computationally efficient and, at the same time, higher frequency would imply that
an episode is occurring often. In our algorithm we use the maximum non-overlapped 
occurences as the frequency measure. This definition of frequency results in very efficient
counting algorithms with some interesting theoretical
properties (\cite{Srivats2005,Srivats-kdd07}).

In analyzing neuronal spike data, it is useful to consider
methods,
where, while counting the frequency, we include only those occurrences which
satisfy some additional temporal constraints. Here we are interested in what we 
call expiry time constraint which is specified by giving a time span $\tau$. The constraint  
 requires that span of occurence of the parallel episode is less that $\tau$.   
 For example in sequence (\ref{eq:data-seq}), with an expiry time $\tau=5$, the occurrence of parallel episode 
 \{${(A,1),(B,3),(C,6)}$\} is valid where as the occurrence \{$(B,3), (C,6), (A,10)$\} is not.
As is easy to see, a parallel episode with expiry time constraints corresponds 
to what we called a {\em synchronous pattern} in the previous section. These are 
the temporal patterns of interest in this paper. To represent a parallel episode 
$(A ~ B ~ C)$ with expiry time $\tau$ we use the notation  $(A~B~C)_\tau$.

Efficient algorithm to count such episodes exist (\cite{PSU2008}). The algorithm counts the non-overlapped 
occurrence of an episode (say, $(A~B~C)_\tau$ ) as follows. 
While going down the data stream it remembers the latest time of occurrence of all its constituent events. 
Once all the events are seen at least once, it checks if the span of the latest occurrences of all the events is less than $\tau$.
If the expiry time constraint is satisfied, then frequency counter is incremented and all the events are marked as not seen.
The algorithm then proceeds further to look for more occurrences.
It is easy to see that such  a method counts only non-overlapped occurrences of the parallel episodes.

However, an efficient counting algorithms alone is not sufficient. This is because at higher levels 
the number of parallel episodes to be counted increases exponentially. The problem of exploding number of 
candidates is tackled through the classic apriori method that is popular in datamining. At each level, the number of 
parallel episodes that have to counted is generated from the frequent candidates at lower levels. 

 Based on this 
idea, we have the following structure for the algorithm. 
 We first get frequent 1-node episodes which are then used to make candidate 2-node 
episodes. Then, by one more pass over data, we find frequent 2-node episodes which are 
then used to make candidate 3-node episodes and so on.  
Such a technique is quite effective in controlling combinatorial explosion and 
 the number of candidates comes down drastically as the size increases. This is because, 
as the size increases, many of the combinatorially possible parallel episodes of that 
size would not be frequent.  This allows 
the algorithm to find large size frequent episodes efficiently. 
 At each stage of this process, we count frequencies 
of not one but a whole set of candidate episodes 
(of a given size) through one sequential pass over 
the data. We do not actually traverse the time axis in time ticks once for each 
pattern  whose occurrences we want to count. We traverse the 
time-ordered data stream. As we traverse the data we remember enough from the data stream to 
correctly take care of all the occurrence possibilities of all episodes in the candidate set 
and thus compute all the frequent episodes of a given size through one pass over the data.  
The complete details of the algorithm are available in (\cite{PSU2008}). 

\section{Significance of discovered synchronous firing patterns}
\label{sec:significance}
In the previous section we discussed effective algorithms to discover synchronous patterns. 
Here, we present significance tests to show that the obtained patterns are significant and 
have not occurred by chance.

There have been many approaches for assessing the significance of detected firing 
patterns (\cite{Abeles2001,Pipa2008}). 
In the current analytical approaches, one generally employs a Null hypothesis 
that the different spike trains are generated by independent processes. 
In many cases one also assumes (possibly inhomogeneous) Bernoulli or  
Poisson processes. Then one can calculate the probability of observing the given 
number of repetitions of the pattern (or of any other statistic derived from such 
 counts) under the null hypothesis of independent processes and hence calculate 
a minimum number of repetitions needed to conclude that a pattern is significant 
 in the sense of being able to reject the null hypothesis. There are also  
 some  empirical approaches, which may be called the jitter methods, suggested for assessing significance. 
Here one creates many surrogate data streams from the 
experimentally observed data by perturbing (or jittering) the individual spikes while keeping 
certain statistics same.  Then, by calculating the empirical distribution of pattern counts on the 
sample of surrogate data, one assesses the significance of the observed patterns.  

 Recently, significance tests for sequential patterns 
had been developed \cite{SU08}. These tests involve estimating the expected frequency of 
serial episodes under a given null hypothesis by modelling the counting process of the algorithm.
We also take a similar approach and modify the method to suit synchronous patterns.
The Null hypothesis we assume is that all the neurons fire independently of each other. 

\subsection{Modelling the counting process}

Suppose, we are operating at a time resolution of $\Delta T$. (That is, 
 the times of events 
or spikes are recorded to a resolution of $\Delta T$). 
Then we discretize the time axis into intervals of length $\Delta T$.
For a parallel episode with expiry times, the span of any occurrence should be less than the expiry time, $T$ 
(in steps of $\Delta T$). But
initially let us assume that the span of each occurrence of a parallel episode is exactly $T$ time steps.
(We can modify the counting to skip $T$ time units once an occurrence is found.)
Now the counting process explained in the previous section can be viewed in the following way. 
 For each episode (say $(A~B~C)_T$) whose 
frequency we want to find, we do the following. We start at time instant 1. We check to see whether there is an 
occurrence of the episode starting from the current instant. If we find an occurrence we increment the
counter and move ahead by $T$ steps and again start looking for another occurrence. If we do not find an occurrence
we move ahead one step. We do this since we reach the end of the data stream of length $L$.

Let $p$ be the probability that we find an occurence of an episode at any given time instant. Then from the 
above description of counting we can say that, from any given time instant we move ahead by $T$ time units with
a probability $p$ and move ahead by one time unit with probability $1-p$. This leads to the recurrence relation,

\begin{equation}
F(L,T,p) = (1-p) F(L-1, T, p) + p ( 1 + F(L-T, L, p))
\label{eq:rec1}
\end{equation}

where, $F(L,T,p)$ is the expected frequency of a episode. The notation $F(L,T,p)$ denotes that
the mean is a function of $L$,$T$ and $p$.

The boundary conditions for this recurrence are:
\begin{equation}
F(x,y,p) = 0, \ \ \mbox{if} \ \ x < y \ \ \mbox{and} \ \  \forall p.
\label{eq:bd-cn}
\end{equation}

Similarly, the mean of the square of the frequency, $G(L,T,p)$, of the episode can be obtained as
\begin{equation}
\begin{split}
\footnotesize
G(L,T,p) = (1-p) G(L-1, T,p) \: + \: \\
p(1 \: + \: G(L-T, T,p) \: + \: 2 F(L-T, T,p))
\label{eq:var-rec1}
\end{split}
\end{equation}

Hence, the variance $V(L,T,p)$ is,

\begin{equation}
V(L,T,p) = G(L,T,p) \: - \: (F(L,T,p))^2 
\label{eq:var}
\end{equation}

Since the neurons fire independently of each other the probability of an occurence of an episode at 
any time instant is given by,

\begin{equation}
p = \rho^n (\Delta T)^n \sum_{i=0}^{n-1} (T-1)^{n-1-i} T^{i} 
\label{eq:p-ind} 
\end{equation}

where, $\rho$ be the unconditional probability 
that a neuron fires at any given time instant. 
We obtain $\rho$ by estimating 
the average rate of firing for this neuron from the data (or we may know it from other prior knowledge). 
Using the value of $p$, we can calculate values of $F(L,T,p)$ and $V(L,T,p)$ from equations 
\ref{eq:rec1}, \ref{eq:var-rec1} and \ref{eq:var}. 
For a given type-I error $\epsilon$, using the Chebyshev inequality, the frequency threshold can then be obtained as $F(L,T,p) + k \sqrt{V(L,T,p)}$, 
where $k$ is the smallest integer such that $k^2 \geq \frac{1}{\epsilon}$.
We use this frequency threshold for mining significant parallel episodes (synchronous firing patterns).

By using this frquency threshold, we ensure that the chances of a random episode being reported as frequent
is less than $\epsilon$. So for low values of $\epsilon$, we can confidentally say that the patterns
reported as frequent are not random patterns.

\section{Simulation Results}
\label{sec:simulations}
In this section we compare 
the parallel episode mining algorithm with a popular existing tool, NeuroXidence, in terms of 
running times, scalability and false positive rates.
NeuroXidence (\cite{Pipa2008}) is used to detect an excess or a lack of synchronous firing in spike train data.
This is done by counting the number of occurrences of synchronous firing patterns satisfying a given expiry time.
Unlike our non-overlapped counts, NeuroXidence counts all occurrences of a pattern. For example, for the pattern, $(A~B~C)_T$ 
any set of spikes of $A$,$B$ and $C$ that satisfy the time constraint is considered as an occurrence. NeuroXidence counts
the frequency of all patterns that occur at least once in the data. The counting process is essentially a correlation based technique. 

NeuroXidence employs a non-parametric method to assess the significance of the observed counts.
The Null hypothesis is that the patterns occur by chance. The estimate of the chance frequency under null hypothesis is
obtained by generating surrogate data. Surrogate data is created by jittering the spikes of the neurons independently of one another. This way the temporal cross structure in the data is destroyed while retaining the auto-structure of the 
spike trains.
For every trial of the data obtained, around 25 surrogates are created. The patterns frequencies are found out in the surrogate data set. 
From the values so obtained, we get an empirical distribution of the chance frequencies. Using that the significance of the 
observed frequency counts are obtained.

NeuroXidence is found to be very effective in finding synchronous firing patterns \cite{Pipa2008}.

\begin{table}[t]
\centering
 \begin{tabular}{|c|c|c|c|c|l|}  \hline
L & \multicolumn{2}{c|}{Avg. Run Time (s)}& \multicolumn{2}{c|}{F.P.R.}  \\ \cline{2-5}
	& PE & NX & PE & NX  \\ \hline
50000		&  0.2 & 51 & 15\% & 31\% \\ \hline
100000 		& 0.375 &134 &21\% &47\% \\ \hline
200000		& 0.8 & 270&48\% &79\% \\ \hline
 \end{tabular}
\caption{Comparison of NeuroXidence (NX) and Parallel Episode Mining Algorithm (PE) : Average running time (in seconds) and False Positive Rates (F.P.R.)comparison 
for varying data lengths ($L$) . (Parameters: $\rho$ = 5 Hz, $T=5$, Number of Neurons = 20.)}
\label{tab:length}
\end{table}

For the results provided in this section we use spike train data generated by using 
the Poisson simulator described in \cite{SU08}.
Each neuron is modelled as an inhomogenous poisson process. Strong interactions among neurons

 can be input to the simulator by means of 
conditional probabilities. For example, if we want the spiking of $A$ at any time $t$ to affect
 the spiking of $B$ at time $t+\tau$,
we represent it by a conditional probality $P(B/(A,\tau)) = p$. 
Since our null hypothesis is of independence no strong connections are embedded into the simulator.
 The neurons fire independently of one another. 
We include correlated firing in the data by means of external stimulation, that is, we embed synchronous firing in the data
generated by the simulator.
Different sized patterns (upto 7 nodes) with various expiry times are embedded in the data.

Effectiveness of both the methods are assessed with respect to the running times and false positive rates.
The methods are tested for varying parameters like random firing rate, different expiry times, 
different number of neurons. The running times and false positives 
rate reported for the parallel episode algorithm are average values obtained from 100 realizations of the data. 
In case of NeuroXidence, the values are averaged over 20 iterations.
The results are reported in Tables~\ref{tab:length}-\ref{tab:exptimes}.

NeuroXidence requires input data from various trials. 
For our experiments we split a single long data into 20 portions and give it as an input.
For better statistical analysis, the number of surrogates for determing the empirical probability distribution is set at 25.

Both the methods were found to be very effective in mining the embedded patterns. 
All the patterns that are embedded in that data were discovered by the both the methods.
However, the parallel episode mining algorithm has huge computational advantage over NeuroXidence (refer Table~\ref{tab:length}) 
Such difference in times are because the NeuroXidence calculates the frequencies of all possible patterns in the data. 
But the parallel episode mining algorithms uses an efficient level wise procedure to count candidates
 generated out of frequent sub-episodes.
Also, the statistical test required for NeuroXidence requires it to find the frequencies of pattern in the surrogate data.
If the number of surrogates is  25, then effectively NeuroXidence has to calculate the frequencies in data that is 25 times longer than 
the input data. This is the reason for the marked difference in running times of the algorithms.

\begin{table}[t]
\centering
 \begin{tabular}{|c|c|c|c|c|l|}  \hline
$\rho$ & \multicolumn{2}{c|}{Avg. Run Time (s)}& \multicolumn{2}{c|}{F.P.R.}  \\ \cline{2-5}
	& PE & NX & PE & NX  \\ \hline
5		&  0.38 & 51 & 22\% & 31\% \\ \hline
10 		& 0.50 &309 &22\% &49\% \\ \hline
 \end{tabular}
\caption{Comparison of NeuroXidence (NX) and Parallel Episode Mining Algorithm (PE) : Average running time (in seconds) and False Positive Rates (F.P.R.)comparison 
for varying random firing frequency ($\rho$) . (Parameters: $L$ = 50000, $T=5$, Number of Neurons = 20.)}
\label{tab:frequency}
\end{table}

The change in expiry time of mining does not affect the running times of the episodes mining
 algorithms(see Table~\ref{tab:exptimes}). 

\begin{table}[h]
\centering
 \begin{tabular}{|c|c|c|c|c|l|}  \hline
$M$ & \multicolumn{2}{c|}{Avg. Run Time (s)}& \multicolumn{2}{c|}{F.P.R.}  \\ \cline{2-5}
	& PE & NX & PE & NX  \\ \hline
20		&  0.38 & 51 & 22\% & 31\% \\ \hline
30 		& 0.44 &233 &27\% &49\% \\ \hline
40		& 0.54 & 1193&40\% &59\% \\ \hline
 \end{tabular}
\caption{Comparison of NeuroXidence (NX) and Parallel Episode Mining Algorithm (PE) : Average running time (in seconds) and False Positive Rates (F.P.R.)comparison 
for varying number of participation neurons ($M$). (Parameters: $\rho$ = 5 Hz, $T=5$, $L = 50000$)}
\label{tab:numberofneurons}
\end{table}

\begin{table}[h]
\centering
 \begin{tabular}{|c|c|c|c|c|l|}  \hline
$T$ 		& \multicolumn{2}{c|}{Avg. Run Time (s)}& \multicolumn{2}{c|}{F.P.R.}  \\ \cline{2-5}
			& PE & NX & PE & NX  \\ \hline
3			&  0.38 & 21 & 29\% & 23\% \\ \hline
5 		& 0.38 &51 &22\% &31\% \\ \hline
8			& 0.37 & 122&15\% &51\% \\ \hline
10		& 0.37 & 189&14\% &54\% \\ \hline
 \end{tabular}
\caption{Comparison of NeuroXidence (NX) and Parallel Episode Mining Algorithm (PE) : Average running time (in seconds) and False Positive Rates (F.P.R.)comparison 
for varying expiry times ($T$) . (Parameters: $\rho$ = 5 Hz, $L=50000$, Number of Neurons = 20.)}
\label{tab:exptimes}
\end{table}

The running time of NeuroXidence increases drastically with expiry times.
 Similar effects can also be seen because of increase in background firing rates(see Table~\ref{tab:frequency}).
The number of false positives increase because more and more patterns start occurring more than once.

The increase in number of neurons has a huge effect on the running times of NeuroXidence(see Table~\ref{tab:numberofneurons}). 
Infact for a network with 40 neurons the running time is as high as 20 minutes for only 50 sec data at 5 Hz. This is because of the exponential increase in the number of patterns to be counted.

From Tables~\ref{tab:frequency} and \ref{tab:exptimes}, it is clear that change in random firing frequency and expiry times
 does not affect the running times very much. The algorithm will be able to scale up for longer data with many interacting neurons.
 
\section{Conclusions}
\label{sec:conclusions}
Fast decoding of information-bearing patterns are critical to the
success of brain-computer interfaces. Data mining approaches, combined
with statistical significance tests that does not require a huge
amount of surrogate data, may provide some of the answers. In this
paper we have presented an approach that significantly more efficient
than existing methods and should lay the foundations for more
efficient decoding of neural signals and hence achieve better
brain-computer interfaces.

\section{Acknowledgements}
Unnikrishnan's work was supported in part by NIH grant U54DA021519.

\end{document}